\documentclass[runningheads]{llncs}
\usepackage[T1]{fontenc}
\usepackage{graphicx}
\usepackage{cite}
\usepackage{amsmath,amssymb,amsfonts}
\usepackage{algorithmic}
\usepackage{algorithm}
\usepackage{graphicx}
\usepackage{textcomp}
\usepackage{xcolor}
\usepackage{hyperref}
\usepackage{graphicx}
\usepackage{amsmath}
\usepackage{amssymb}
\usepackage{multirow}      
\usepackage{multicol}
\usepackage{subfigure}
\usepackage{booktabs}
\usepackage{array}
\usepackage{makecell}
\usepackage{threeparttable}
\usepackage{bm}
\usepackage{float}
\newcommand{\etal}{\textit{et al.}}
\newcommand{\ie}{\textit{i.e.}}
\begin{document}
\title{Weight Prediction Boosts the Convergence of AdamW}
%
%
\author{Lei Guan}
\authorrunning{Lei Guan}
%
\institute{Department of Mathematics, National University of Defense Technology
\email{guanleimath@163.com}
}
\maketitle              
\begin{abstract}
In this paper, we introduce weight prediction into the AdamW optimizer to boost its convergence when training the deep neural network (DNN) models. In particular, ahead of each mini-batch training, we predict the future weights according to the update rule of AdamW and then apply the predicted future weights to do both forward pass and backward propagation. In this way, the AdamW optimizer always utilizes the gradients w.r.t. the future weights instead of current weights to update the DNN parameters, making the AdamW optimizer achieve better convergence. Our proposal is simple and straightforward to implement but effective in boosting the convergence of DNN training. We performed extensive experimental evaluations on image classification and language modeling tasks to verify the effectiveness of our proposal. The experimental results validate that our proposal can boost the convergence of AdamW and achieve better accuracy than AdamW when training the DNN models.

\keywords{Deep learning  \and Weight prediction  \and Convergence \and AdamW.}
\end{abstract}
\section{Introduction}
The optimization of deep neural network (DNN) models is to find the optimal parameters using an optimizer which has a decisive influence on the convergence of the models and thus directly affects the total training time.  Adaptive gradient methods, such as RMSprop~\cite{tieleman2012lecture}, AdaGrad~\cite{duchi2011adaptive}, Adam~\cite{kingma2014adam} and AdamW~\cite{loshchilov2017decoupled}, are currently of core practical importance in deep learning training as they are able to attain rapid training of modern deep neural network (DNN) models.  Particularly, AdamW~\cite{loshchilov2017decoupled}, also known as Adam with decoupled weight decay, has been used as a default optimizer for training various DNN models~\cite{tieleman2012lecture,loshchilov2017decoupled,liu2021swin,bai2021transformers,wang2022scaled}. The major advantage of AdamW lies in that it improves the generalization performance of Adam~\cite{kingma2014adam} and thus works as effectively as SGD with momentum~\cite{sutskever2013importance} on image classification tasks.

As with other popular gradient-based optimization methods, when using AdamW as an optimizer, each iteration of DNN training, \ie, \ a mini-batch training, generally consists of one forward pass and one backward propagation, where the gradients w.r.t. all the parameters (also known as weights) are computed during the backward propagation. The generated gradients are then used by the AdamW optimizer to calculate the update values for all parameters, which are finally applied to updating the weights. The remarkable features of using AdamW to update parameters include: 1)  the updates of weights are continuous; 2) each mini-batch uses the currently available weights to do both forward pass and backward propagation. 

Motivated by the fact that DNN weights are updated in a continuous manner and the update values calculated by the AdamW should reflect the "correct" direction for updating the weights, we introduce weight prediction~\cite{chen2018efficient, guan2019xpipe} into the DNN training to further boost the convergence of AdamW. Concretely, ahead of each mini-batch training, we first perform weight prediction according to the currently available weights and the update rule of AdamW. Following that, we use the predicted future weights instead of current weights to perform both forward pass and backward propagation. Finally, the AdamW optimizer utilizes the gradients w.r.t. the predicted weights to update the DNN parameters. We experiment with two typical machine learning tasks, including image classification and language modeling. The experimental results demonstrate that our proposal outperforms AdamW in terms of convergence and accuracy. For instance, when training four convolution neural network (CNN) models on CIFAR-10 dataset, our proposal yields an average accuracy improvement of 0.47\% (up to 0.74\%) over AdamW. When training LSTMs on Penn TreeBank dataset, our proposal achieves 5.52 less perplexity than AdamW on average.

The contributions of this paper can be summarized as follows:
\begin{itemize} 
	\item [(1)] We, for the first time, construct the mathematical relationship between currently available weights and future weights after several continuous updates  when using AdamW as an optimizer.
	\item [(2)] We devise an effective way to incorporate weight prediction into AdamW. To the best of our knowledge, this is the first time that uses weight prediction strategy to boost the convergence of AdamW.  The proposed weight prediction strategy is believed to be well suited for other popular optimization methods such as RMSprop~\cite{tieleman2012lecture}, AdaGrad~\cite{duchi2011adaptive}, Adam~\cite{kingma2014adam}, et al.
	\item [(3)] We conducted extensive experimental evaluations to validate the effectiveness of our proposal, which demonstrates that our proposal is able to boost the convergence of AdamW when training the DNN models.
\end{itemize}


\section{Related Work}
When using the gradient-based optimization methods to train DNN models, the differences in optimization methods lie in that the ways using gradients to update model parameters are different. Generally, the commonly used first-order gradient methods can be categorized into two groups: the accelerated stochastic gradient descent (SGD) family~\cite{nesterov27method,sutskever2013importance,polyak1964some} and adaptive gradient methods~\cite{zeiler2012adadelta,kingma2014adam,zhuang2020adabelief}.

Adaptive gradient methods, also known as adaptive learning methods, have been heavily studied in prior research and widely used in deep learning training. Very different from the SGD methods (e.g., Momentum SGD~\cite{sutskever2013importance}), which use a unified learning rate for all parameters, adaptive gradient methods compute a specific learning rate for each individual parameter~\cite{zhuang2020adabelief}. In 2011, Duchi et al.~\cite{duchi2011adaptive} proposed the AdaGrad, which can dynamically adjust the learning rate according to the history gradients from previous iterations and utilize the quadratic sum of all previous gradients to update the model parameters. Zeiler~\cite{zeiler2012adadelta} proposed AdaDelta, seeking to alleviate the continual decay of the learning rate of AdaGrad. AdaDelta does not require manual tuning of a learning rate and is robust to noisy gradient information. Tieleman and Hinton~\cite{tieleman2012lecture} refined AdaGrad and proposed RMSprop. The same as AdaGrad, RMSprop adjusts the learning rate via element-wise computation and then updates the variables. One remarkable feature of RMSprop is that it can avoid decaying the learning rate too quickly. In order to combine the advantages of both AdaGrad and RMSprop, Kingma and Ba~\cite{kingma2014adam} proposed another famous adaptive gradient method, Adam, which has become an extremely important choice for deep learning training. Loshchilov and Hutter~\cite{loshchilov2017decoupled} found that the major factor of the poor generalization of Adam is due to that $L_2$ regularization for it is not as effective as for its competitor, the Momentum SGD. They thus proposed decoupled weight decay regularization for Adam, which is also known as AdamW. The experimental results demonstrate that AdamW substantially improves the generalization performance of Adam and illustrates competitive performance as Momentum SGD~\cite{sutskever2013importance} when tackling image classification tasks. To simultaneously achieve fast convergence and good generalization, Zhuang \etal~\cite{zhuang2020adabelief} proposed another adaptive gradient method called AdaBelief, which adapts the stepsize according to the ``belief'' in the current gradient direction. Other adaptive gradient methods include AdaBound~\cite{luo2019adaptive}, RAdam~\cite{liu2019variance}, Yogi~\cite{zaheer2018adaptive}, et al. It is worth noting that all these adaptive gradient methods share a common feature: weight updates are continuous and each mini-batch training always uses currently available weights to perform both forward pass and backward propagation.

Weight prediction was previously used to overcome the weight inconsistency issue in the asynchronous pipeline parallelism. Chen et al.~\cite{chen2018efficient} used the smoothed gradient to replace the true gradient in order to predict future weights when using Momentum SGD~\cite{sutskever2013importance} as the optimizer. Guan et al.~\cite{guan2019xpipe} proposed using the update values of Adam~\cite{kingma2014adam} to make weight predictions.  Yet, both approaches use weight prediction to ensure the weight consistency of pipeline training rather than considering the impact of weight prediction on the optimizers themselves.

\section{Methods}
Ahead of any $t$-th ($t\geq1$) iteration, we assume that the current available DNN weights are ${\bm \theta}_{t-1}$. Given the initial learning rate $\gamma\in \mathbb{R}$, momentum factor $\beta_1 \in \mathbb{R}$ and $\beta_2 \in \mathbb{R}$, and weight decay value $\lambda \in \mathbb{R}$, we reformulate the update of AdamW~\cite{loshchilov2017decoupled} as
\begin{equation}
	\begin{aligned}
		& {\bm \theta}_{t} = (1-\gamma\lambda){\bm \theta} _{t-1} - \frac{\gamma{\hat{\bf m}_t}}{\sqrt{\hat{\bf v}_t}+\epsilon}, \\
		\textnormal{s.t.} 
		& \left\{\begin{array}{ll}
			\mathbf{g}_t = \nabla_{t}f({\bm \theta}_{t-1}),\\
			{\mathbf m}_{t}= \beta_1\cdot \mathbf m_{t-1} + (1-\beta_1) \cdot \mathbf{g}_t,\\
			\mathbf{v}_t = \beta_2\cdot \mathbf{v}_{t-1} + (1-\beta_2)\cdot \mathbf{g}_t^2, \\
			\hat{\mathbf m}_{t}= \frac{\mathbf m_{t}}{1-\beta_1^t},\\
			\hat{\mathbf v}_{t}= \frac{\mathbf v_t}{1-\beta_2^t}.
		\end{array}\right.
	\end{aligned}
	\label{adamw_update}
\end{equation}
In~\eqref{adamw_update}, ${\mathbf m}_{t}$ and ${\mathbf v}_{t}$ refer to the first and second moment vector respectively, $\epsilon$ is the smoothing term which can prevent division by zero.

Letting ${\bm \theta}_0$ denote the initial weights of a DNN model, then in the following $s$ times of continuous mini-batch training, the DNN weights are updated  via
\begin{equation}
	\begin{aligned}
		& {\bm \theta}_1 = (1-\gamma\lambda){\bm \theta}_0 - \frac{\gamma{\hat{\bf m}_1}}{\sqrt{\hat{\bf v}_1}+\epsilon}, \\
		& {\bm \theta}_2 = (1-\gamma\lambda){\bm \theta}_1 - \frac{\gamma{\hat{\bf m}_2}}{\sqrt{\hat{\bf v}_2}+\epsilon}, \\
		& \cdots \\
		& {\bm \theta}_s = (1-\gamma\lambda){\bm \theta}_{s-1} - \frac{\gamma{\hat{\bf m}_s}}{\sqrt{\hat{\bf v}_s}+\epsilon},\\
	\end{aligned}
\label{adamw_update_series}
\end{equation}
where for any $ i \in\{1,2,\cdots, s\}$, we have
\begin{equation}
	\begin{aligned}
		\left\{\begin{array}{ll}
			\mathbf{g}_i = \nabla_{i}f({\bm \theta}_{i-1}), \\
			{\mathbf m}_{i}= \beta_1\cdot \mathbf m_{i-1} + (1-\beta_1) \cdot \mathbf{g}_i,\\
			\mathbf{v}_i = \beta_2\cdot \mathbf{v}_{i-1} + (1-\beta_2)\cdot \mathbf{g}_i^2, \\
			\hat{\mathbf m}_{i}= \frac{\mathbf m_{i}}{1-\beta_1^i},\\
			\hat{\mathbf v}_{i}= \frac{\mathbf v_i}{1-\beta_2^i}. \\
		\end{array}\right.
	\end{aligned}
\end{equation}

When summing up all weight update equations in~\eqref{adamw_update_series}, we have
\begin{equation}
	\begin{aligned}
	& {\bm \theta}_s = {\bm \theta}_0 - \gamma\lambda\sum_{i=0}^{t-1}({\bm \theta}_0+{\bm \theta}_1+\cdots+{\bm \theta}_{t-1}) - \sum_{i=1}^t\frac{\gamma \hat{\bf m}_i}{\sqrt{\hat{\bf v}_i}+\epsilon},\\
	& \textnormal{s.t.}
	\left\{\begin{array}{ll}
		\mathbf{g}_i = \nabla_{i}f({\bm \theta}_{i-1}), \\
		{\mathbf m}_{i}= \beta_1\cdot \mathbf m_{i-1} + (1-\beta_1) \cdot \mathbf{g}_i,\\
		\mathbf{v}_i = \beta_2\cdot \mathbf{v}_{i-1} + (1-\beta_2)\cdot \mathbf{g}_i^2, \\
		\hat{\mathbf m}_{i}= \frac{\mathbf m_{i}}{1-\beta_1^i},\\
		\hat{\mathbf v}_{i}= \frac{\mathbf v_i}{1-\beta_2^i}. \\
	\end{array}\right.
	\end{aligned}
\label{adam_sum}
\end{equation}

It is well known that the weight decay value $\lambda$ is generally set to an extremely small value (e.g., $5e^{-4}$), and the learning rate $\gamma$ is commonly set to a value smaller than 1 (e.g., 0.01). Consequently, $\gamma\lambda$ is pretty close to zero, and thus, the second term of the right hand of~\eqref{adam_sum} can be neglected. This, therefore, generates the following equation:
\begin{equation}
	{\bm \theta}_s \approx {\bm \theta}_0  - \sum_{i=1}^s\frac{\gamma \hat{\bf m}_i}{\sqrt{\hat{\bf v}_i}+\epsilon}.
	\label{adam_sum_approx}
\end{equation}

\eqref{adam_sum_approx} illustrates that given the initial weights ${\bm \theta}_0$, the weights after $s$ times of continuous updates can be approximately calculated. Correspondingly, given ${\bm \theta}_t$, the weights after $s$ times of continuous updates can be approximately calculated via
\begin{equation}
	\bm \theta_{t+s} \approx \bm \theta_{t} - \sum_{i=t+1}^{t+s}\frac{\gamma \hat{\bf m}_i}{\sqrt{\hat{\bf v}_i}+\epsilon}.
	\label{theta_step}
\end{equation}

From~\eqref{theta_step}, we see that given the initial weights $\bm \theta_t$, $\bm \theta_{t+s}$ can be approximately calculated by letting $\bm \theta_t$ subtract the sum of $s$ continuous relative variation of the weights. Note that the relative increments of the weights in each iteration should reflect the trend of the weight updates in each iteration.  In \eqref{theta_step}, $\frac{\gamma \hat{\bf m}_i}{\sqrt{\hat{\bf v}_i}+\epsilon}$  should reflect the ``correct'' direction for updating the weights ${\bm \theta}_t$ as it is calculated by the AdamW, and the weights are updated in a continuous manner and along the way of inertia directions. 


We can therefore replace $\sum_{i=t+1}^{t+s}\frac{\gamma \hat{\bf m}_i}{\sqrt{\hat{\bf v}_i}+\epsilon}$ in~\eqref{theta_step} with $s\frac{\gamma \hat{\bf m}_{t+1}}{\sqrt{\hat{\bf v}_{t+1}}+\epsilon}$ in an effort to approximately predict $\bm \theta_{t+s}$ for the case when only $\bm \theta_t$, $ \mathbf{g}_t$ and the weight prediction steps $s$ are available. Note that at any $t$-th iteration, the gradients of stochastic objective, \ie, \ $\mathbf{g}_t = \nabla_t (\bm{\theta}_{t-1})$, can be calculated when the backward propagation is completed. Letting ${\hat {\bm \theta}_{t+s}}$ denote the approximately predicted weights for ${\bm \theta}_{t+s}$, we can construct the mathematical relationship between ${\bm \theta} _t$ and $\hat{\bm \theta} _{t+s}$ as
\begin{equation}
	\begin{aligned}
	&\hat{\bm \theta}_{t+s} = \bm \theta_t - s\frac{\gamma \hat{\bf m}_{t+1}}{\sqrt{\hat{\bf v}_{t+1}}+\epsilon}, \\
	\textnormal{s.t.}
	&\left\{\begin{array}{ll}
		\mathbf{g}_t = \nabla_{t}f({\bm \theta}_{t-1}), \\
		{\mathbf m}_{t}= \beta_1\cdot \mathbf m_{t-1} + (1-\beta_1) \cdot \mathbf{g}_t,\\
		\mathbf{v}_t = \beta_2\cdot \mathbf{v}_{t-1} + (1-\beta_2)\cdot \mathbf{g}_t^2, \\
		\hat{\mathbf m}_{t}= \frac{\mathbf m_{t}}{1-\beta_1^t},\\
		\hat{\mathbf v}_{t}= \frac{\mathbf v_t}{1-\beta_2^t}. \\
	\end{array}\right.
	\end{aligned}
	\label{weight_update}
\end{equation}

In the following, we showcase how to incorporate weight prediction into the DNN training when using AdamW~\cite{loshchilov2017decoupled} as an optimizer. Algorithm \ref{alg1} illustrates the detailed information. The weight prediction step $s$ and other hyper-parameters are required ahead of the DNN training. At each iteration, the current  available weights ${\bm \theta}_t$ should be cached before the forward pass starts (Line 4).  Then weight prediction are performed using ~\eqref{weight_update} and the predicted weights  $\hat{\bm \theta}_{t+s}$ is generated (Line 5). Following that, the predicted weights  $\hat{\bm \theta}_{t+s}$ are used to do both forward pass and backward propagation (Lines 6 and 7). Finally, the cached weights ${\bm \theta}_t$ is recovered and updated using the AdamW optimizer (Lines 8 and 9).


\begin{algorithm}[htb]
	\centering
	\caption{Weight prediction for AdamW}
	\label{alg1}
	\begin{algorithmic}[1]
		\REQUIRE Weight prediction step $s$, other hyper-parameters such as $\gamma$,  $\beta_1$, $\beta_2$, $\gamma$, $\epsilon$.
		\STATE {Initialize or load DNN weights $\bm \theta_0$.}
		\STATE {$t \leftarrow{1}$.}
		\WHILE {stopping criterion is not met}
		\STATE{Cache the current weights $\bm \theta_t$.}
		\STATE{Calculate $\hat{\bm \theta}_{t+s}$ using \eqref{weight_update}.}
		\STATE {Do forward pass with $\hat{\bm \theta}_{t+s}$.}
		\STATE {Do backward propagation with $\hat{\bm \theta}_{t+s}$.}
		\STATE {Recover the cached weights $\bm \theta_t$.}
		\STATE {Update the weights $\bm \theta_t$ using the AdamW optimizer. }
		\STATE{$t\leftarrow t+1$.}
		\ENDWHILE
	\end{algorithmic}
\end{algorithm}

\section{Experiments}
\subsection{Experiment Settings}
In this section, we mainly compare our proposal with AdamW~\cite{loshchilov2017decoupled}. We evaluated our proposal with three different weight prediction steps (\ie, $s=1$, $s=2$, and $s=3$), which were respectively denoted as Ours-S1, Ours-S2, and Ours-S3 for convenience purposes.  We conducted experimental evaluations on two different machine learning tasks: image classification on the CIFAR-10~\cite{krizhevsky2009learning} dataset with four CNN models and language modeling on Penn TreeBank~\cite{marcinkiewicz1994building} dataset with two LSTM~\cite{ma2015long} models. 
All the experiments were conducted on a multi-GPU platform which is equipped with four NVIDIA Tesla P100 GPUs, each with 16GB of memory size. The CPU on the platform is Intel Xeon E5-2680 with 128GB DDR4-2400 off-chip main memory.

The CIFAR-10 dataset totally includes 60k 32$\times$32 images, 50k images for training, and 10k images for validation. The Penn TreeBank dataset  consists of 929k training words, 73k validation words as well as 82k test words. For image classification, the used CNN models are VGG-11~\cite{simonyan2014very}, ResNet-34~\cite{he2016deep}, DenseNet-121~\cite{huang2017densely}, and Inception-V3~\cite{szegedy2016rethinking}. For language modeling, we trained the LSTM models with two sizes: 1-layer LSTM and 2-layer LSTM. Each layer was configured with 650 units and was applied 50\% dropout on the non-recurrent connections.

We trained all CNN models for 120 epochs with a mini-batch size of 128. The learning rate was initialized as $1e^{-4}$, and divided by ten at the 90th epoch. For training 1-layer and 2-layer LSTM models, we set the size of  each mini-batch to 20.  We trained both LSTM models for 100 epochs with an initial learning rate of 0.01 and decreased the learning rate by a factor of 10 at the 60th and 80th epochs. For AdamW~\cite{loshchilov2017decoupled} and our proposal, we always evaluated them with the default parameters, \ie, $\beta_1=0.9$, $\beta_2=0.999$, and $\epsilon=10^{-8}$. The weight decay for both approaches was set to $\lambda=5e^{-4}$.





\subsection{CNNs on CIFAR-10}
\begin{figure*}[!h]
	\centering
	\subfigure[VGG-11]{\includegraphics[width=.48\textwidth]{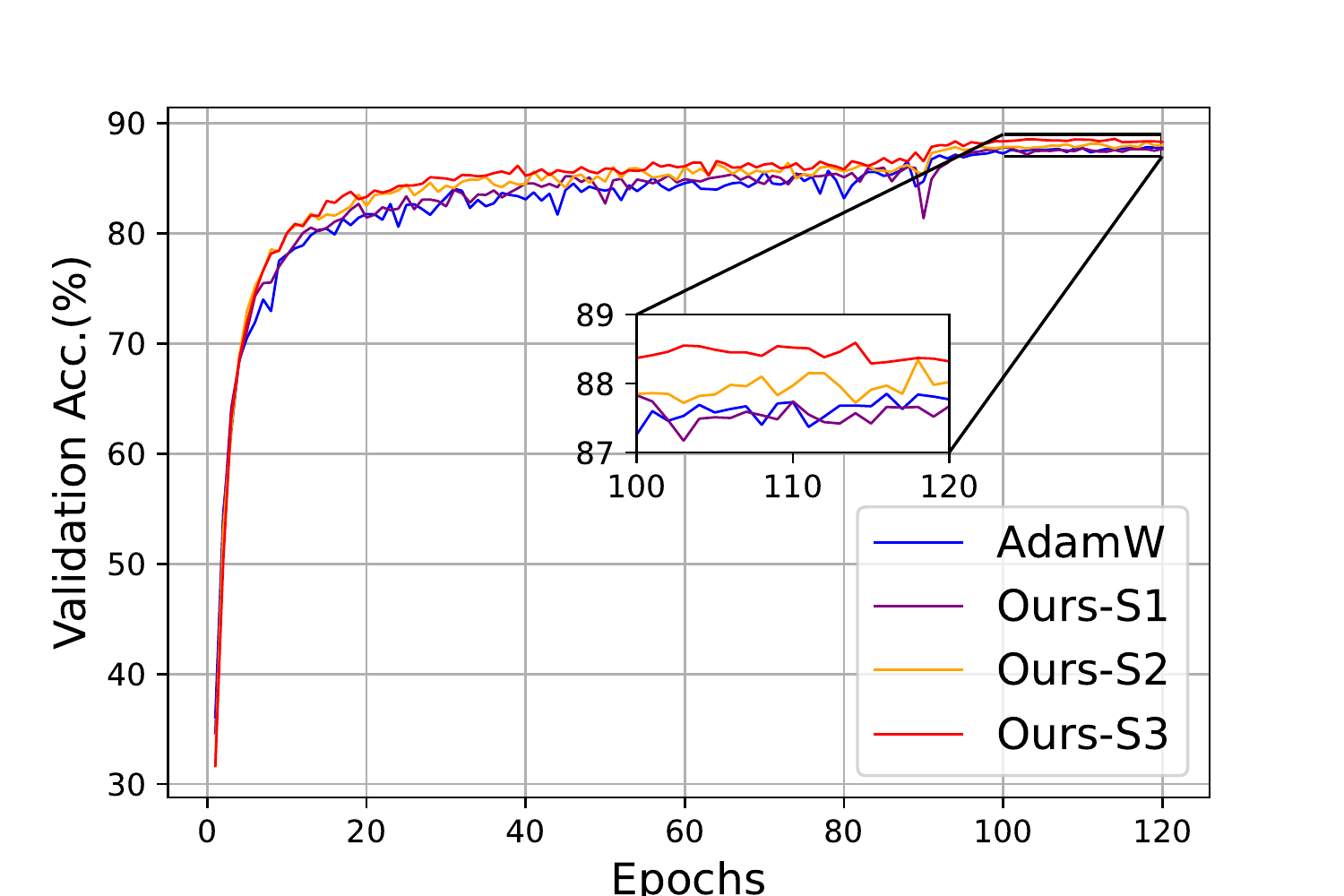}\label{comp-vgg16-acc}}
	\subfigure[ResNet-34]{\includegraphics[width=.48\textwidth]{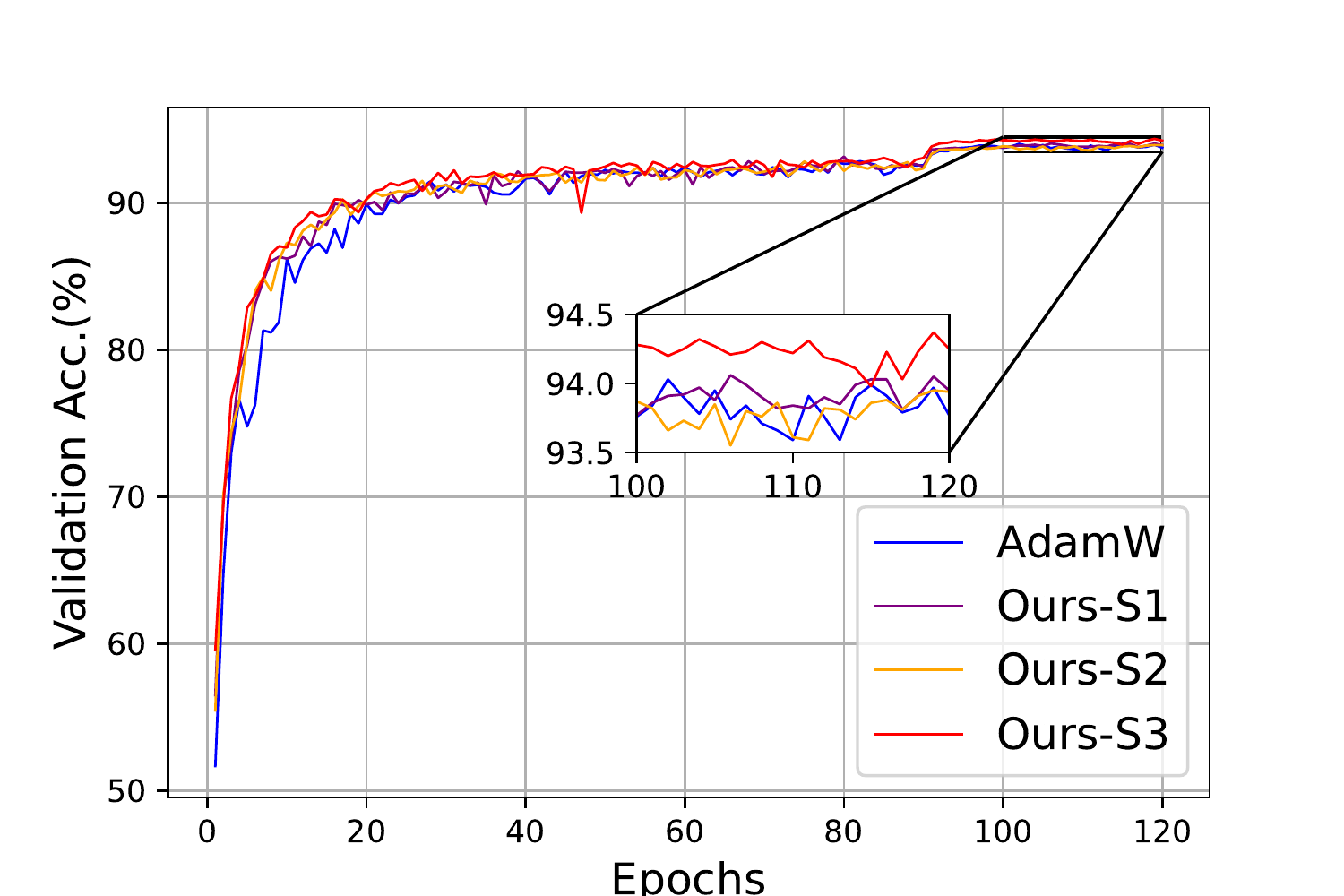}\label{comp-resnet34-acc}}
	\subfigure[DenseNet-121]{\includegraphics[width=.48\textwidth]{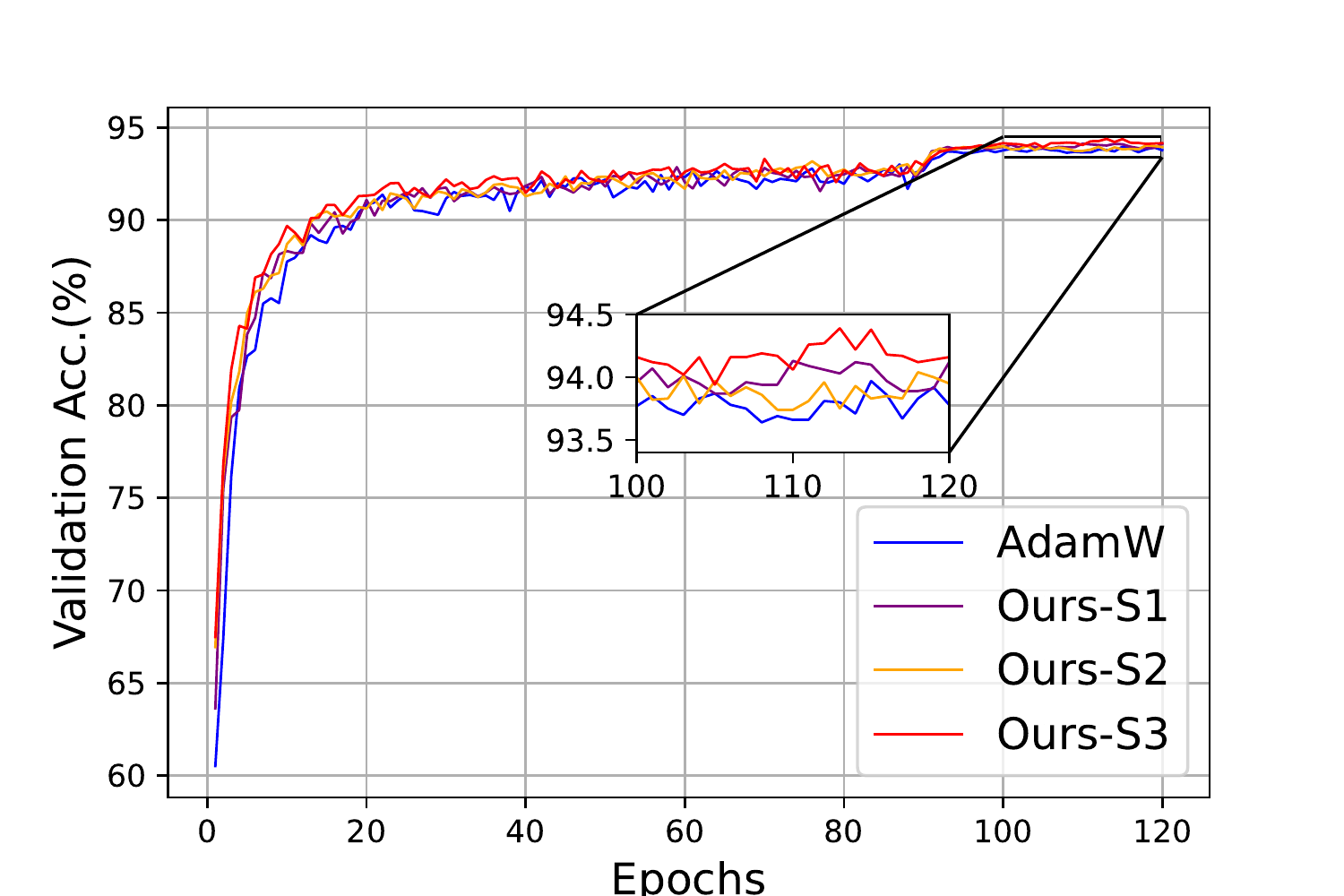}\label{comp-densenet-acc}}
	\subfigure[Inception-V3]{\includegraphics[width=.48\textwidth]{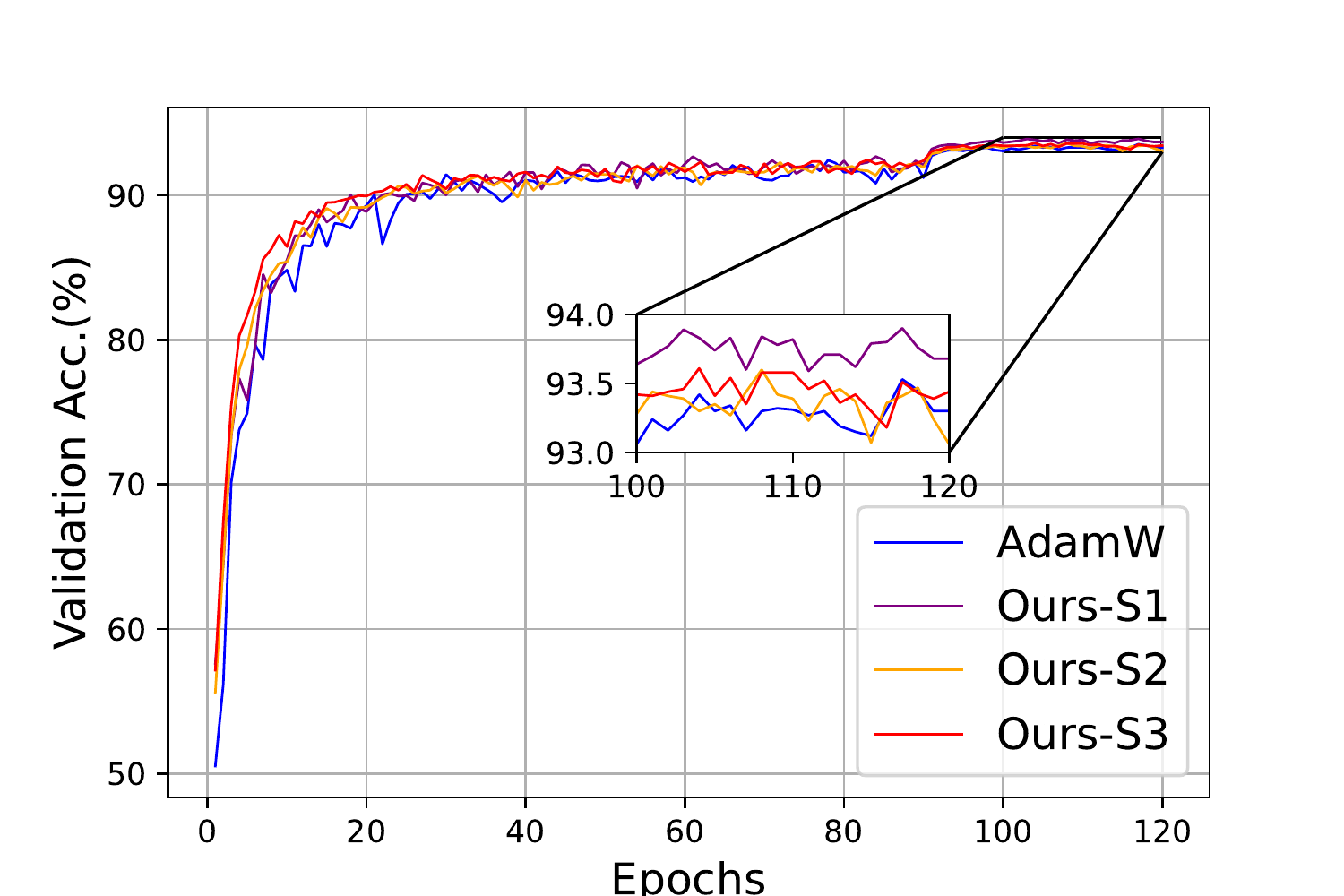}\label{comp-inception-acc}}
	\caption{Validation accuracy vs. epochs of training VGG-11, ResNet-34, DenseNet-121 and Inception-V3 on CIFAR-10.}
	\label{comp-acc-cifar10}
\end{figure*}

In this section, we report the experimental results when training four CNN models on the CIFAR-10 dataset. Table~\ref{table:cifar-acc} summarizes the maximum validation top-1 accuracy and Table~\ref{table:cifar-loss} presents the minimum validation loss. Figure~\ref{comp-acc-cifar10} depicts the learning curves of validation accuracy vs. epochs for training CNNs using AdamW, Ours-S1, Ours-S2, and  Ours-S3, respectively. The learning curves about validation loss vs. epochs are shown in Figure~\ref{table:cifar-loss}.

\begin{table*}[!h]
	\centering
	\caption{Maximum validation top-1 accuracy on CIFAR-10. \textbf{Higher} is better.}
	\label{table:cifar-acc}
	\setlength{\tabcolsep}{3mm}
	\begin{tabular}{c|cccc}
		\toprule
		Models & AdamW &  Ours-S1   &Ours-S2   & Ours-S3    \\
		\midrule
		VGG-11 & 87.85\% & 87.83\% & 88.34\% & \textbf{88.59}\% \\
		ResNet-34 & 94.03\%  & 94.06\% & 93.95\% &  \textbf{94.37}\% \\
		DenseNet-121& 93.97\% & 94.13\% &  94.04\% &  \textbf{94.39}\%\\
		Inception-V3 & 93.53\% & \textbf{93.90}\% & 93.60\% &  93.61\%\\
		\bottomrule
	\end{tabular}
\end{table*}

\begin{figure*}[!h]
	\centering
	\subfigure[VGG-11]{\includegraphics[width=.48\textwidth]{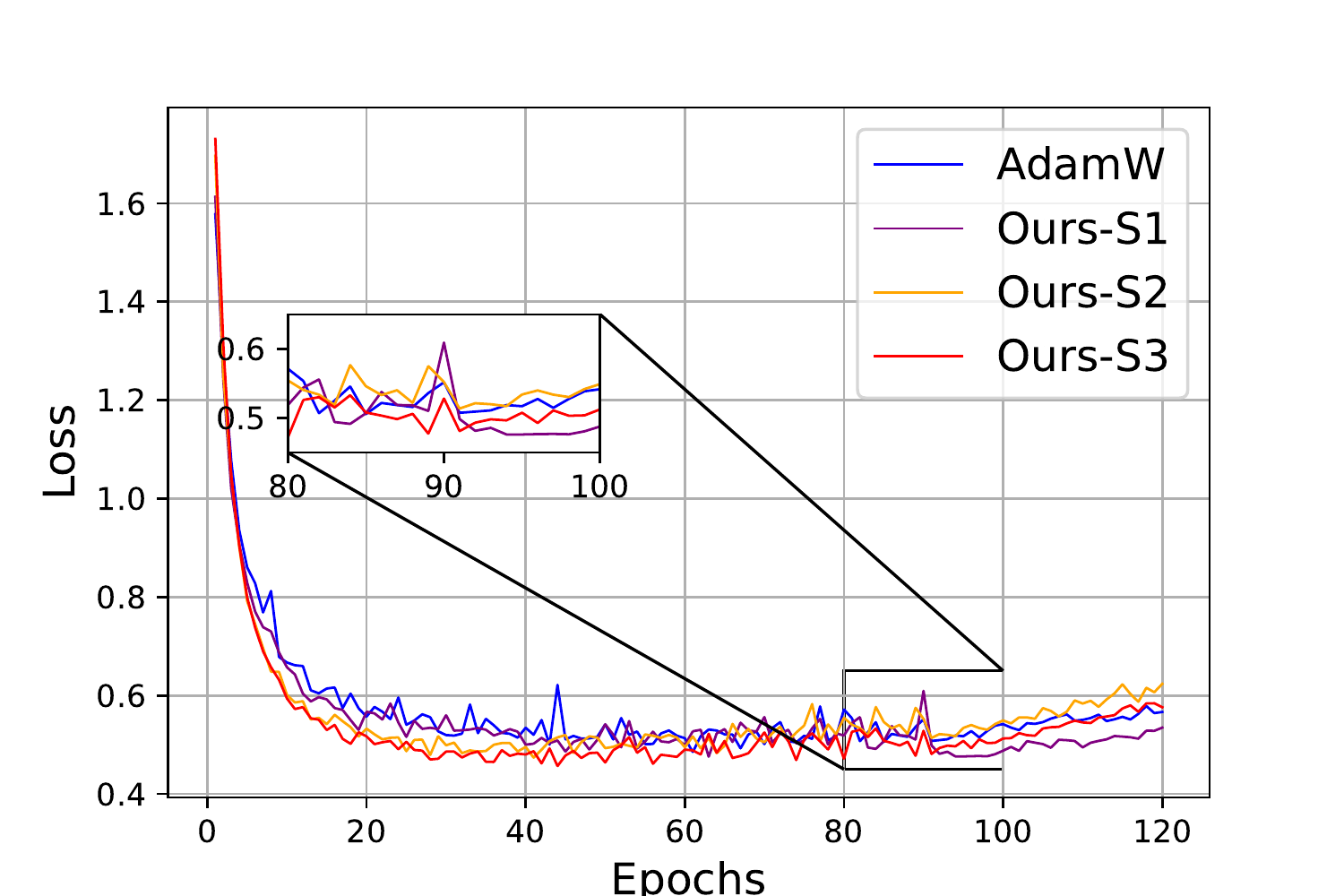}\label{comp-vgg16-loss}}
	\subfigure[ResNet-34]{\includegraphics[width=.48\textwidth]{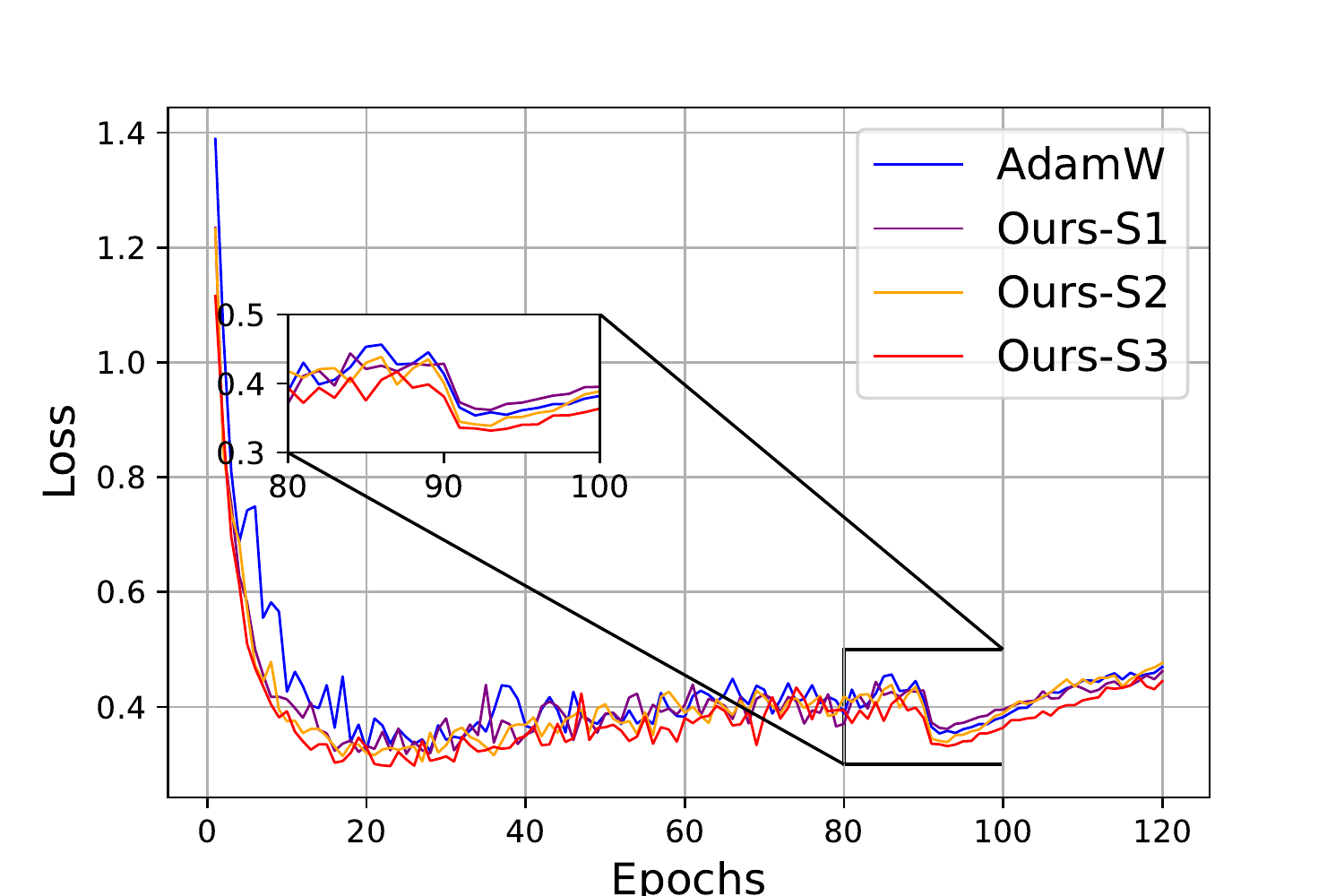}\label{comp-resnet34-loss}}
	\subfigure[DenseNet-121]{\includegraphics[width=.48\textwidth]{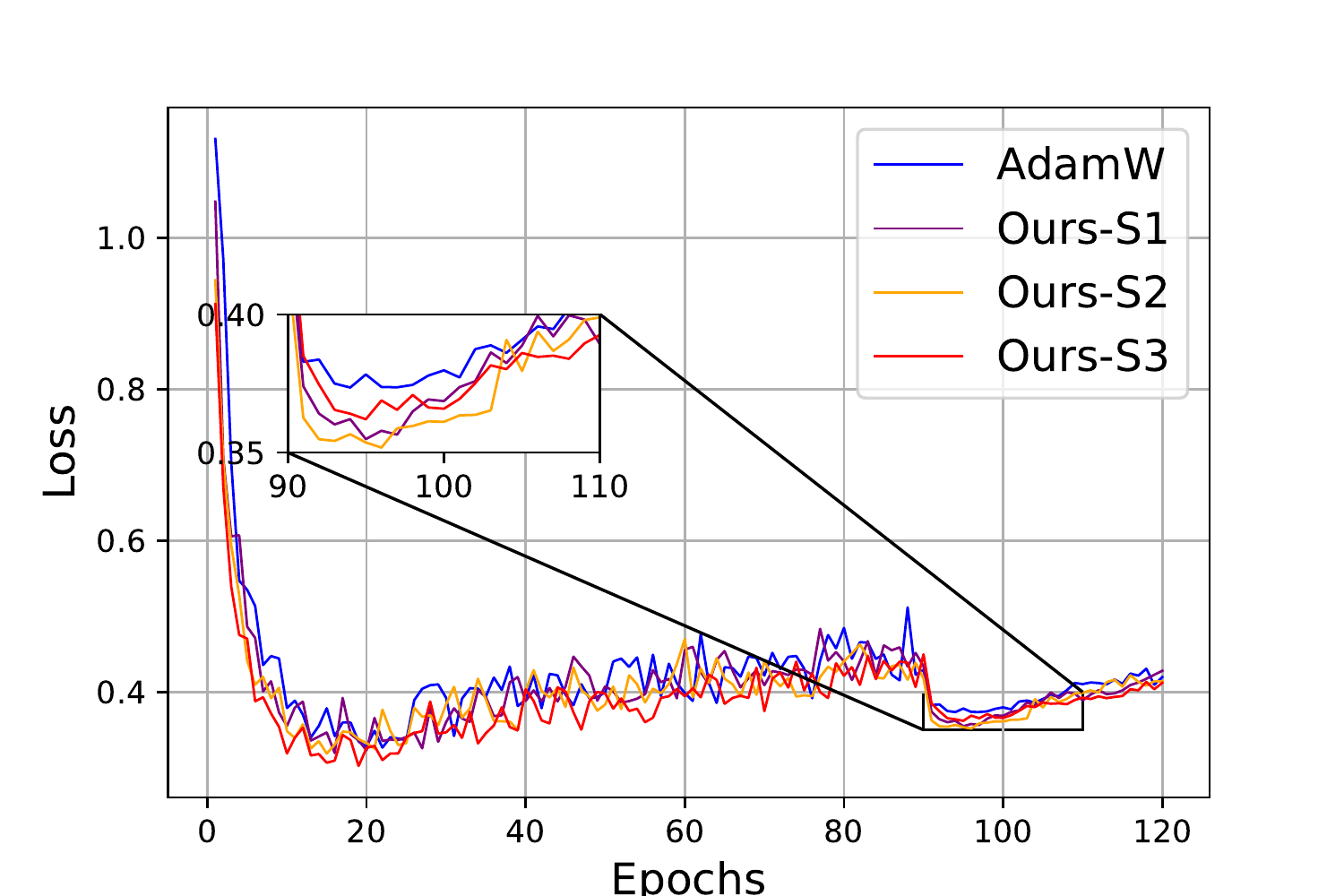}\label{comp-densenet-loss}}
	\subfigure[Inception-V3]{\includegraphics[width=.48\textwidth]{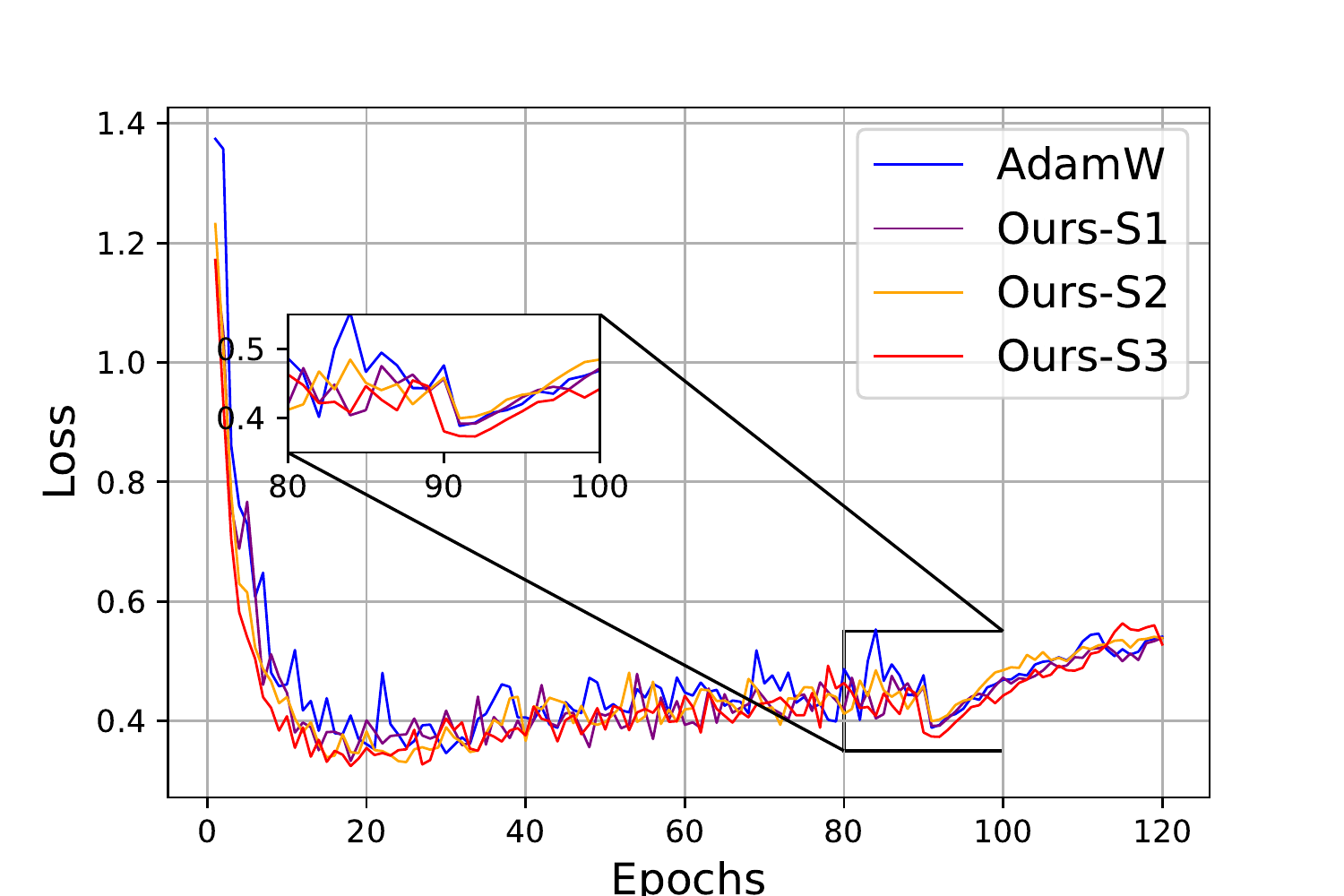}\label{comp-inception-loss}}
	\caption{Validation loss vs. epochs of training VGG-11, ResNet-34, DenseNet-121 and Inception-V3 on CIFAR-10.}
	\label{comp-loss-cifar10}
\end{figure*}

\begin{table*}[!h]
	\centering
	\caption{Minimum validation loss on CIFAR-10. \textbf{Lower} is better.}
	\label{table:cifar-loss}
	\setlength{\tabcolsep}{3mm}
	\begin{tabular}{c|cccc}
		\toprule
		Models & AdamW & Ours-S1   &Ours-S2   & Ours-S3   \\
		\midrule
		VGG-11 & 0.485& 0.475& 0.474& \textbf{0.456} \\
		ResNet-34 & 0.323 & 0.318 & 0.305 &  \textbf{0.297} \\
		DenseNet-121& 0.327 &  0.319&  0.319 &  \textbf{0.302} \\
		Inception-V3 & 0.346 & 0.333 & 0.331 &  \textbf{0.324}\\
		\bottomrule
	\end{tabular}
\end{table*}

Based on the observation of Table~\ref{table:cifar-acc} and Figure~\ref{comp-acc-cifar10}, we can immediately reach the following conclusions. First, Figure~\ref{comp-acc-cifar10} shows that the learning curves of our proposal with different weight prediction steps match well with that of AdamW but converge faster than that of AdamW, especially at the beginning of training epochs. The learning curves in Figures~\ref{comp-vgg16-acc}, \ref{comp-resnet34-acc}, \ref{comp-densenet-acc}, and~\ref{comp-inception-acc} also illustrate that our proposal generally attains higher validation accuracy than AdamW at the end of the training. Second, Table~\ref{table:cifar-acc} shows that our proposal outperforms AdamW on all evaluated CNN models in terms of the obtained maximum validation accuracy. In particular, our proposal achieves consistently higher validation top-1 accuracy than AdamW. Compared to AdamW, our proposal achieves 0.74\%, 0.34\%, 0.42\%, and 0.37\% for training VGG-11, ResNet-34, DenseNet-121, and Inception-V3, respectively. On average,  our proposal yields 0.47\% (up to 0.74\%) top-1 accuracy improvement over AdamW. Third, comparing the experimental results of Ours-S1, Ours-S2, and Ours-S3, we can see that our proposal with different weight prediction steps consistently gets good results, which demonstrates that the performance of our proposal is independent of the settings of the weight prediction step.  Particularly, the experimental results show that Ours-S3 works the best for VGG-11, ResNet-34, and DenseNet-121, while Ours-S1 works the best for Inception-V3. Similar conclusions can be drawn from the observation of Table~\ref{table:cifar-loss} and Figure~\ref{comp-loss-cifar10}. Our proposal consistently obtains less validation loss than AdamW which again verifies that weight prediction can boost the convergence of AdamW when training DNN models.


\subsection{LSTMs on Penn TreeBank}
\begin{figure*}[!h]
	\centering
	\subfigure[Loss]{\includegraphics[width=.48\textwidth]{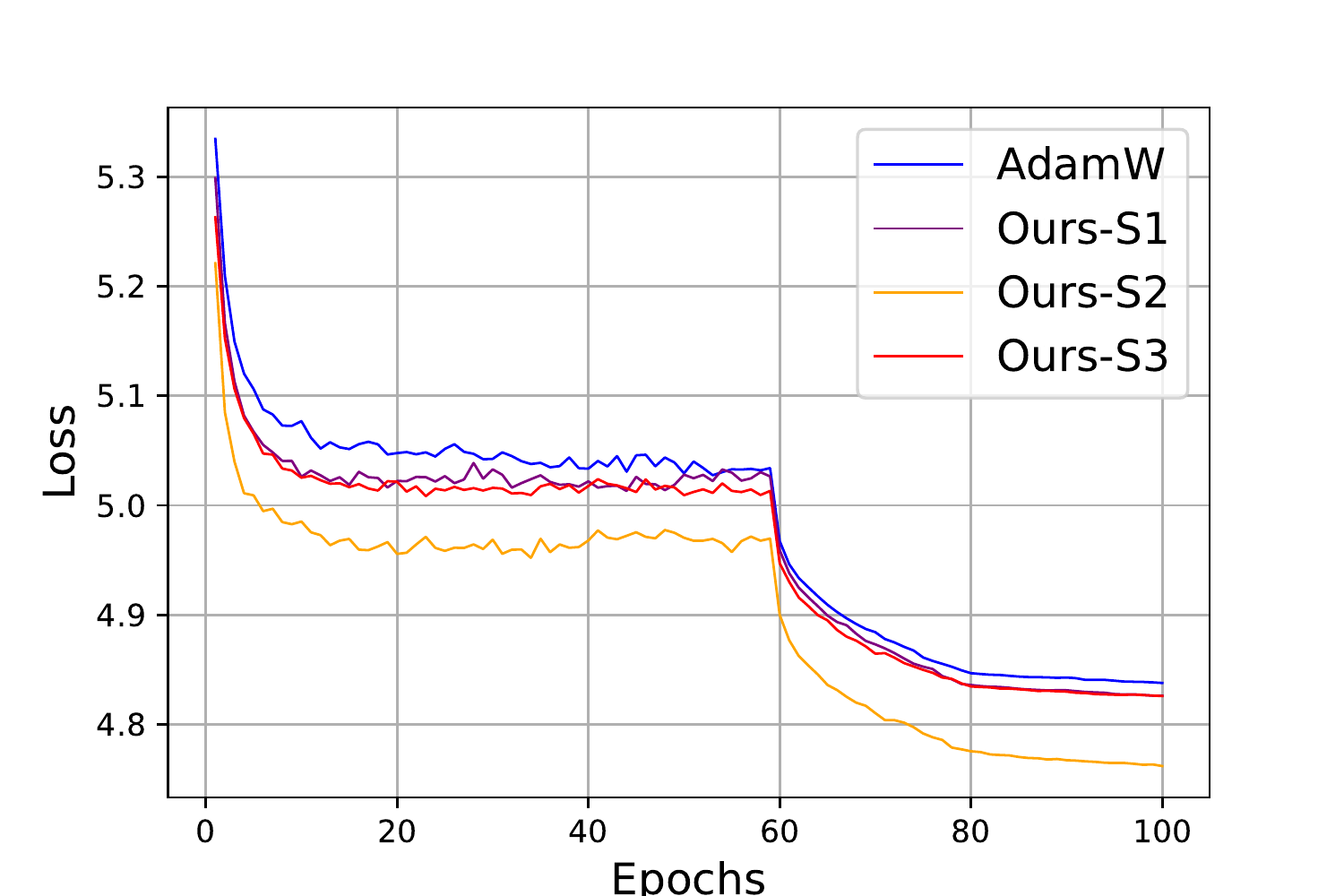}\label{comp-lstm1-loss}}
	\subfigure[Perplexity]{\includegraphics[width=.48\textwidth]{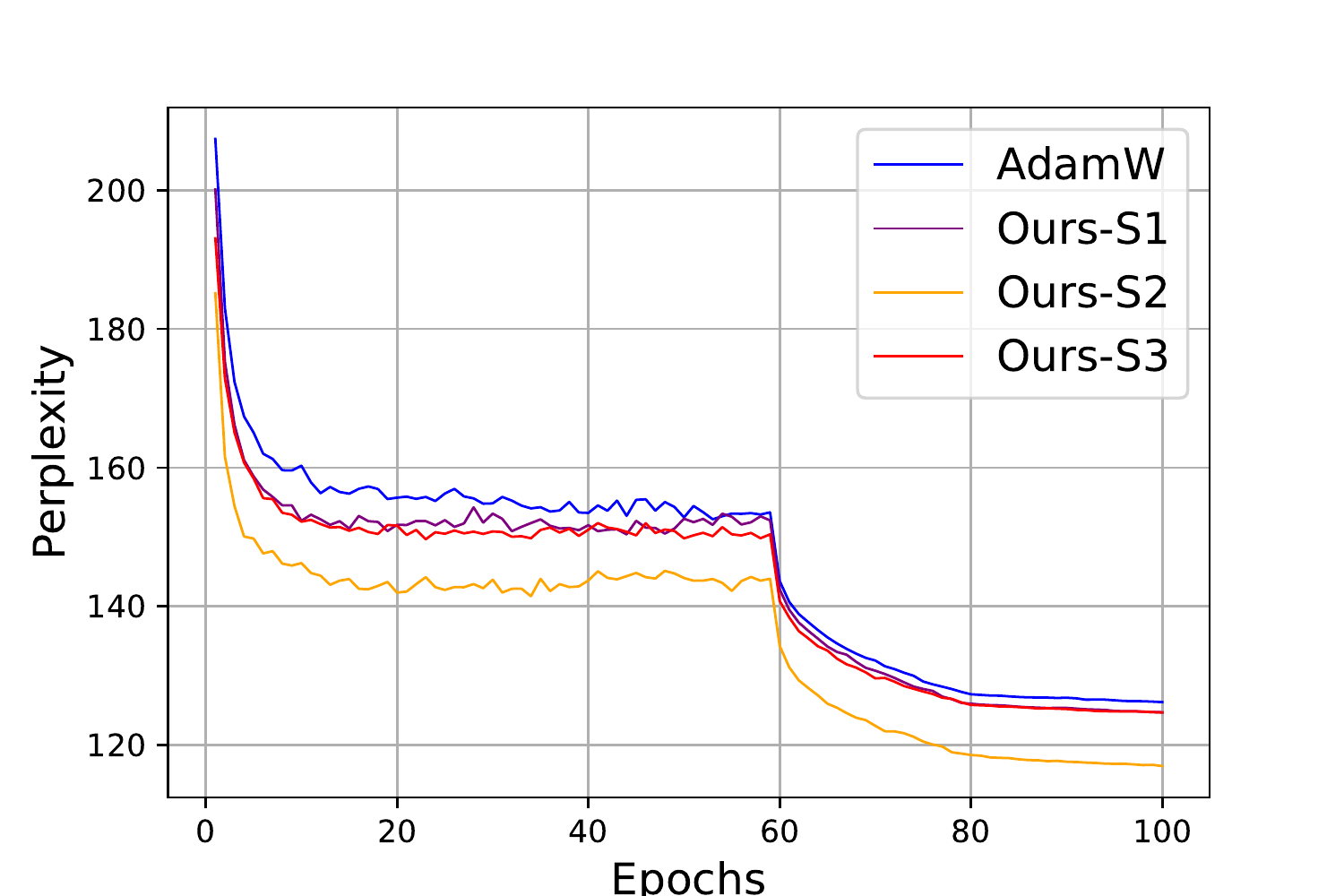}\label{comp-lstm1-ppl}} \qquad
	\caption{Training 1-layer LSTM on Penn TreeBank. Left: Loss vs.  epochs; Right: Perplexity vs. epochs. }
	\label{comp-lstm1}
\end{figure*}

\begin{figure*}[!h]
	\centering
	\subfigure[Loss]{\includegraphics[width=.48\textwidth]{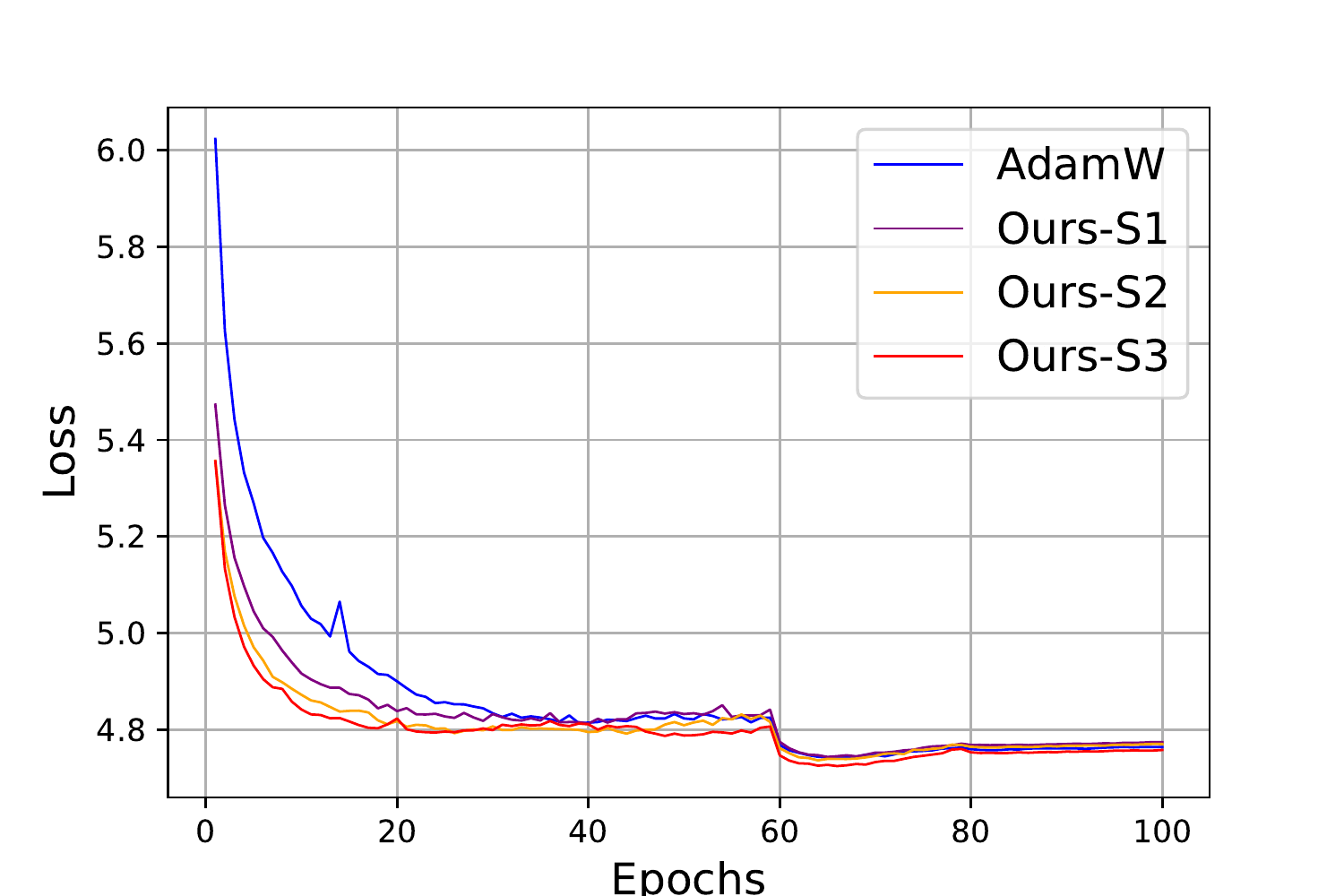}\label{comp-lstm2-loss}}
	\subfigure[Perplexity]{\includegraphics[width=.48\textwidth]{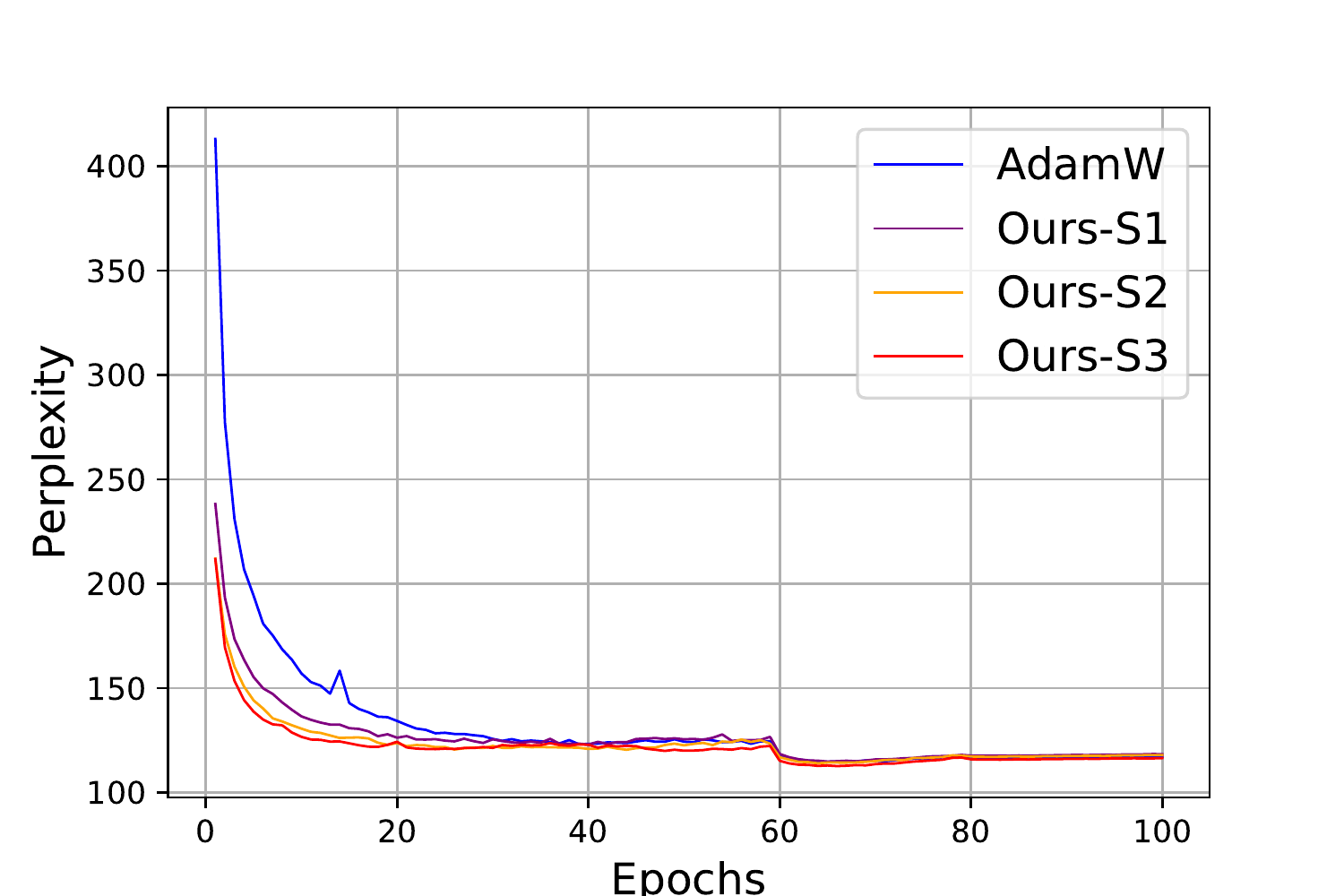}\label{comp-lstm2-ppl}}
	\caption{Training 2-layer LSTM on Penn TreeBank. Left: Loss vs.  epochs; Right: Perplexity vs. epochs.}
	\label{comp-lstm2}
\end{figure*}

In this section, we report the experimental results when training 1-layer and 2-layer LSTM models on the Penn TreeBank dataset~\cite{marcinkiewicz1994building}. Figures~\ref{comp-lstm1} and~\ref{comp-lstm2} depict the learning curves. Table~\ref{table:lstm-ppl} presents the obtained minimum perplexity (lower is better), and Table~\ref{table:lstm-loss} summarizes the obtained minimum validation loss (lower is better).





\begin{table*}[!h]
	\centering
	\caption{Minimum perplexity on Penn TreeBank. \textbf{Lower} is better.}
	\label{table:lstm-ppl}
	\setlength{\tabcolsep}{3mm}
	\begin{tabular}{c|cccc}
		\toprule
		Models & AdamW & Ours-S1   &Ours-S2   & Ours-S3   \\
		\midrule
		1-Layer LSTM& 126.21 & 124.75& \textbf{116.99}&124.71 \\
		2-Layer LSTM& 114.68  &  114.80&  113.99 &  \textbf{112.64} \\
		\bottomrule
	\end{tabular}
\end{table*}

\begin{table*}[!h]
	\centering
	\caption{Minimum validation loss on Penn TreeBank. \textbf{Lower} is better.}
	\label{table:lstm-loss}
	\setlength{\tabcolsep}{3mm}
	\begin{tabular}{c|cccc}
		\toprule
		Models & AdamW & Ours-S1   &Ours-S2   & Ours-S3  \\
		\midrule
		1-Layer LSTM & 4.838 &  4.826& \textbf{4.762}& 4.826 \\
		2-Layer LSTM& 4.742  & 4.743& 4.736 &   \textbf{4.724}\\
		\bottomrule
	\end{tabular}
\end{table*}

We can draw the following conclusions from the experiment results. First, as shown in Table~\ref{table:lstm-ppl}, for both 1-layer and 2-layer LSTM models, our proposal achieves lower perplexity and validation loss than AdamW, validating the fast convergence and good accuracy of our proposal. Second, for 1-layer LSTM, our proposal with $s=2$ yields 9.22 less perplexity than AdamW. For 2-layer LSTM, our proposal with $s=3$ yields 2.02 less perplexity than AdamW. On average, our proposal achieves 5.52 less perplexity than AdamW. Second, similar conclusions can be drawn based on the observation of the loss vs. epochs learning curves in Figures~\ref{comp-lstm1-loss} and~\ref{comp-lstm2-loss} and Table~\ref{table:lstm-loss}. Our proposal consistently achieves less validation loss than AdamW, again validating that weight prediction can boost the convergence of AdamW.


\section{Conclusions}
To further boost the convergence of AdamW, in this paper, we introduce weight prediction into the DNN training.
The remarkable feature of our proposal is that we perform both forward pass and backward propagation using the future weights which are predicted according to the update of AdamW. In particular, we construct the mathematical relationship between current weights and future weights and devise an effective way to incorporate weight prediction into DNN training. Our proposal is easy to implement and works well in boosting the convergence of DNN training. The experimental results on image classification and language modeling tasks verify the effectiveness of our proposal.

The weight prediction should also work well for other adaptive optimization methods such as RMSprop~\cite{tieleman2012lecture}, AdaGrad~\cite{duchi2011adaptive}, and Adam~\cite{kingma2014adam} et al. when training the DNN models. For future work, we would like to apply weight prediction to those popular optimization methods.

\bibliographystyle{splncs04}
 \bibliography{mybibliography}
%




\end{document}